\pdfoutput=1

\documentclass[11pt]{article}

\usepackage[]{ACL2023}

\usepackage{times}
\usepackage{latexsym}

\usepackage[T1]{fontenc}

\usepackage[utf8]{inputenc}

\usepackage{microtype}

\usepackage{inconsolata}

\usepackage{url}
\usepackage{amsfonts} 
\usepackage{booktabs} 
\usepackage{multirow} 
\usepackage{subfigure} 
\usepackage{xspace}
\usepackage{graphicx}
\usepackage{soul}
\usepackage{color,xcolor}
\usepackage{colortbl}
\usepackage[english]{babel}
\usepackage{balance}
\usepackage{amsmath,bm}
\usepackage{makecell}
\usepackage{algorithm}
\usepackage{algpseudocode}
\usepackage{makecell}
\usepackage{float}
\usepackage{CJKutf8}

\usepackage{array}
\newcommand{\PreserveBackslash}[1]{\let\temp=\\#1\let\\=\temp}
\newcolumntype{C}[1]{>{\PreserveBackslash\centering}p{#1}}
\newcolumntype{R}[1]{>{\PreserveBackslash\raggedleft}p{#1}}
\newcolumntype{L}[1]{>{\PreserveBackslash\raggedright}p{#1}}

\definecolor{Gray}{gray}{0.9}

\title{ELLA: Empowering LLMs\\for Interpretable, Accurate  and Informative Legal Advice}

\author{
    Yutong Hu$^{1,2}$\thanks{\quad Equal Contribution.}, 
    Kangcheng Luo$^{3,*}$, 
    \textbf{Yansong Feng}$^{1}$\thanks{\;\;Corresponding author.}~~\\ 
    $^1$Wangxuan Institute of Computer Technology, Peking University, China \\ 
    $^2$School of Intelligence Science and Technology, Peking University \\
    $^3$School of Electronics Engineering and Computer Science, Peking University, China \\ 
    State Key Laboratory of General Artificial Intelligence \\
    {\tt \{huyutong,fengyansong\}}
     {\tt @pku.edu.cn} \\
     {\tt luokangcheng@stu.pku.edu.cn}\\
}

\begin{document}
\maketitle
\begin{abstract}
  Despite remarkable performance in legal consultation exhibited by legal Large Language Models(LLMs) 
  combined with legal article retrieval components, there are still cases when the advice given is incorrect or baseless. 
  To alleviate these problems, we propose {\bf ELLA}, a tool for {\bf E}mpowering {\bf L}LMs for interpretable, accurate, and informative {\bf L}egal {\bf A}dvice. ELLA visually presents the correlation between legal articles and LLM's response by calculating their similarities, providing users with an intuitive legal basis for the responses. Besides, based on the users' queries, ELLA retrieves relevant legal articles and displays them to users. Users can interactively select legal articles for LLM to generate more accurate responses. 
  ELLA also retrieves relevant legal cases for user reference. Our user study shows that presenting the legal basis for the response helps users understand better. The accuracy of LLM's responses also improves when users intervene in selecting legal articles for LLM. Providing relevant legal cases also aids individuals in obtaining comprehensive information. 
  Our github repo is: \url{https://github.com/Huyt00/ELLA}\footnote{Video demonstration is available at: \url{https://youtu.be/V8iaIXSJ2i8}}.

\end{abstract}

\section{Introduction}
\label{sec:intro}
Large Language Models (LLMs), such as LLAMA~\cite{touvron2023llama}, ChatGLM~\cite{zeng2023glm130b} and GPT4~\cite{openai2024gpt4}, have shown impressive performance in various tasks, 
showing great potential for specific domains, such as law~\cite{lai2023large} and finance~\cite{wu2023bloomberggpt, yang2023fingpt}. In the legal domain, many attempts have been made\cite{colombo2024saullm7b,lawyer-llama-report,yue2023disclawllm,nguyen2023brief,cui2023chatlaw}, which acquire legal knowledge through continual training and performing a supervised fine-tuning stage with a large-scale legal dataset. These models can offer various services including legal consultations, explaining legal terminology, analyzing legal cases, and preparing legal documents.

Despite the remarkable performance of LLMs within the legal domain, they are not exempt from the occurrence of hallucination~\cite{Ji_2023}. To alleviate this, previous studies~\cite{lawyer-llama-report,yue2023disclawllm,cui2023chatlaw} have proposed retrieval-augmented generation(RAG)~\cite{lewis2021retrievalaugmented} frameworks to retrieve legal articles from an external datastore. 
By leveraging retrieved legal articles, hallucination is reduced and LLMs can generate more faithful answers. 

\begin{figure}[pt]
    \center
    \includegraphics[width=0.48\textwidth]{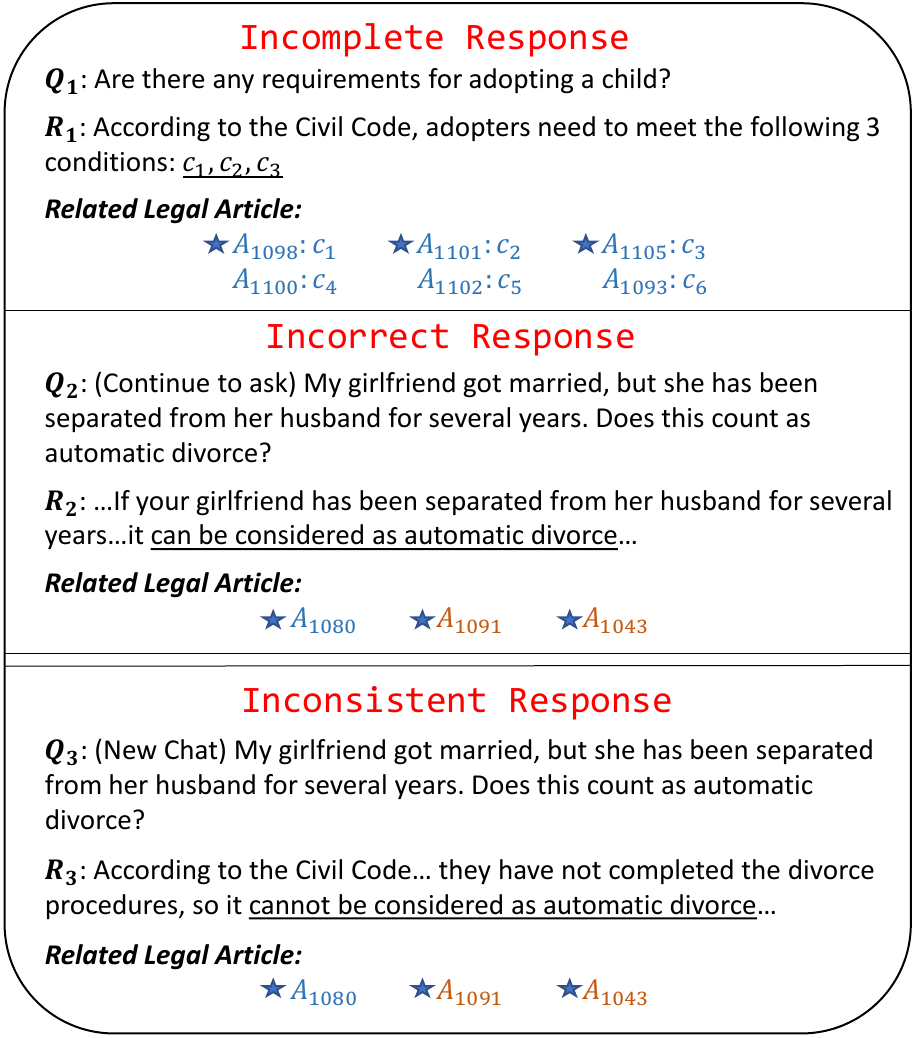}
    \caption{Examples of incomplete, incorrect, inconsistent Response. $A_i$ indicates the $i_{th}$ article in Civil Code. Blue articles mean they are relevant to the query, while orange ones are irrelevant. The blue star means the article is retrieved for LLM. We only show the key information in the Figure. For the complete conversations, please refer to Appendix~\ref{app:conversation}}
    \label{fig:example}
\end{figure}


In the legal domain, LLMs' responses are required to have high accuracy and be supported by reasonable legal bases. Therefore, the retrieval component plays an important role as it provides correct and related legal articles for LLMs. 
While LLMs could be augmented with retrieved legal articles to generate faithful responses, when irrelevant ones are retrieved, they inevitably bring noise to LLMs, leading LLMs to produce responses with incomplete, incorrect or inconsistent information. 

For instance, as shown in Figure~\ref{fig:example}, when a user asks $Q_1$, the legal article retrieval model retrieves articles 1098, 1101, and 1105 of the Civil Code~\footnote{\url{https://www.gov.cn/xinwen/2020-06/01/content_5516649.htm}} according to the query, while fails to retrieve another three relevant ones: article 1100, 1102 and 1093 of the Civil Code. Therefore, LLM only suggests that adopters need to meet the conditions $c_1, c_2$ and $c_3$ mentioned in the retrieved legal articles, resulting in incomplete suggestions. Then the user continues to ask $Q_2$. Although the related article is retrieved, the irrelevant ones are also retrieved. Such irrelevant articles bring noise to LLM, leading to the incorrect response $R_2$(In fact, the case mentioned in $Q_2$ should not be considered as automatic divorce). Besides, LLMs may be sensible to the input perturbation~\cite{zhu2023promptbench,dong2023revisit}. Responses can be contradictory when inputs only differ slightly. For example, when the user begins a new chat and asks $Q_3$, which is identical to $Q_2$, the response $R_3$ is contradictory to $R_2$. This inconsistency can potentially bring confusion to users, resulting in a lower-quality consultation. 

When LLMs fail to produce coherent and complete responses, relevant legal cases can offer users more in-depth reference information~\cite{su2024caseformer}. However, a legal case retrieval module has rarely been integrated into the existing legal domain LLMs in civil law systems. Additionally, legal terminology may sometimes be embedded in the responses lacking sufficient explanations, posing potential understanding difficulties for users without domain knowledge~\cite{savelka2023explaining}.

To address the issues mentioned above, we propose {\bf ELLA}, a tool {\bf E}mpowering {\bf L}LMs for interpretable, accurate, and informative {\bf L}egal {\bf A}dvice. 

Firstly, we fine-tune BGE~\cite{bge_embedding}, an embedding model for retrieval, to retrieve the legal basis for each sentence in the response. By visually presenting the legal basis to users, users can trust the advice provided by LLMs. When there is no legal basis for a sentence, it can be viewed as a warning that the sentence may be incorrect. Secondly, ELLA retrieves several legal articles based on the user's query and presents them to users. Users can interactively select the relevant legal articles for LLMs to generate accurate and complete responses while disregarding irrelevant ones to avoid noise. Thirdly, 
we incorporate a legal case retrieval model in ELLA, intending to present supplementary information for users to reference. Considering the long context in legal cases, we find all the key sentences in the article through similarity matching between the query and each sentence in the legal case. We highlight all key sentences in the legal cases for users to improve their reading efficiency.

The response interpretation aids users in understanding and placing trust in the advice given by LLMs. The user study shows that our model can generate more accurate responses when users interactively select relevant legal articles. The legal case retrieval module also offers users more resourceful reference information.

\section{Framework and Usage Example}
ELLA is composed of four parts: 1) {\bf Chat Interface}: visually displays the conversation between the user and the LLM. 2) {\bf Interactive Legal Article Selection:} Provides retrieved legal articles for users to choose from, letting the LLM generate new responses based on the user's selected legal articles. 3) {\bf Response Interpretation:} Provides legal article and judicial interpretations to interpret each sentence of the LLM's response. 4) {\bf Legal Case Retrieval:} Displays relevant legal cases for the user to refer to.

\begin{figure*}[pt]
    \center
    \includegraphics[width=1\textwidth]{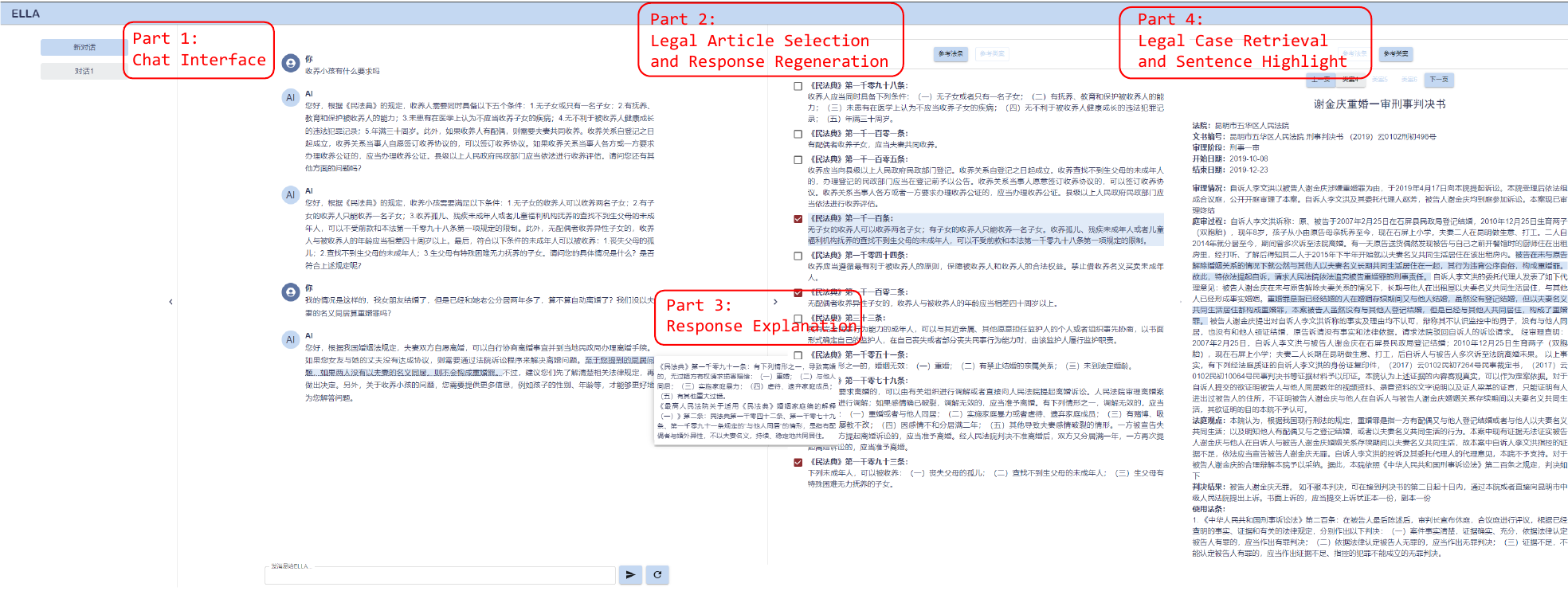}
    \caption{Screenshot of ELLA. We show the complete conversation in Appendix~\ref{app:conversation}, Table~\ref{table:q1} and Table~\ref{table:q4}.}
    \label{fig:frontend}
\end{figure*}

\subsection{Chat Interface} 
\label{sec:chat}
Our chat interface is shown in Figure~\ref{fig:frontend}, part 1.  After clicking the input button, the chat box above will display user input and the LLM's response. Users can have multiple rounds of chats, or click 'new conversation' on the upper left to start a new consultation. The column on the left retains all conversations. Users can click on each chat button to view the corresponding chat content.

\subsection{Interactive Legal Article Selection}
Legal article retrieval model plays an important part in Chinese legal domain LLMs~\cite{lawyer-llama-report,yue2023disclawllm}. Lawyer LLaMA~\cite{lawyer-llama-report} mentions that when LLMs are provided with external relevant legal articles, they can generate more reliable responses. However, the current legal article retrieval models cannot ensure to retrieval all the relevant legal articles and leave out all irrelevant ones. Missed articles might reduce the completeness of the model's response, while irrelevant articles bring noise to LLM, leading LLMs to generate irrelevant advice.

To solve this problem, ELLA allows users to interactively select legal articles. We display the top $K_1=10$ relevant legal articles retrieved for the users. Users can select relevant legal articles based on their situations. The LLM will then generate responses based on the legal articles selected by the user. Note that the LLM generates its first response based on the top 3 retrieved articles by default. Subsequently, users can select legal articles for LLM to regenerate new responses multiple times. 

Back to the example in Figure~\ref{fig:example}, we find that several relevant legal articles are not selected for LLM. Then we can select them, as shown in Figure~\ref{fig:frontend}, part 2, and click the "Regenerate" button at the bottom of the page. Then LLM generates a new response with complete information. By allowing users to participate in the legal article retrieval, it increases the consistency between the user's situation and the referred legal articles used by the LLM, thus enabling the LLM to generate more complete and accurate responses.

\subsection{Response Interpretation}
The response interpretation module provides the legal article basis for each sentence in the LLM's response, and helps users better understand the terminologies in the responses. 

LLM is sensitive to the inputs. Users may receive different advice when ask the same questions in different ways. To facilitate users to identify which response is more reliable, or whether a response is trustworthy, the response interpretation module presents the referred legal articles for each sentence in the response. Users can verify the reliability of the response by tracing the legal article basis of each sentence. 

At the same time, even though the LLM can conveniently provide legal advice to users, sometimes the responses may contain terminologies, which non-professional users may find hard to understand. Besides, some special cases lack a clear definition in the legal articles. They are both explicitly explained in China's "judicial interpretations". To provide users with a better legal consultation experience, we use a response explanation module to provide a clear explanation of the terminology/special cases with corresponding judicial interpretation, making it easier for users to understand.

As shown in Figure~\ref{fig:frontend}, part 3, when the user ask {\it "My girlfriend is married...Would living with her without being legally married be considered bigamy?"}, the response is {\it "...The situation you mentioned is cohabitation rather than bigamy...cohabitation is not illegal..."}. To check the definition of "cohabitation", the user can hover the mouse over the sentence. Then the platform will display a hovering box, showing the corresponding judicial interpretation. We show the legal article basis for the sentence in the same way. If there is neither a legal article basis nor a judicial interpretation for the sentence, the hovering box will not display.

\subsection{Legal Case Retrieval}
Legal cases also serve as important references for users when they consult on legal issues and make judgments about their circumstances. Currently, Chinese legal domain LLMs can only make decisions for users based on internal legal knowledge and externally retrieved legal articles, unable to provide relevant legal cases for users as reference. Therefore, we introduced a legal case retrieval module in ELLA. For every query from users, we search relevant legal cases obtained from China Judgements Online~\footnote{\url{https://wenshu.court.gov.cn}} and display them on the platform for users, as shown in Figure~\ref{fig:frontend}, part 4. As the context of the legal cases may be long, we highlight the sentences in the trial proceeding records related to the user's query. Users can directly locate these sentences to get key information. We provide multiple relevant legal cases. Users can click the button at the top of part 4 to view different legal cases.

\section{System Overview}

\begin{figure*}[pt]
    \center
    \includegraphics[width=1\textwidth]{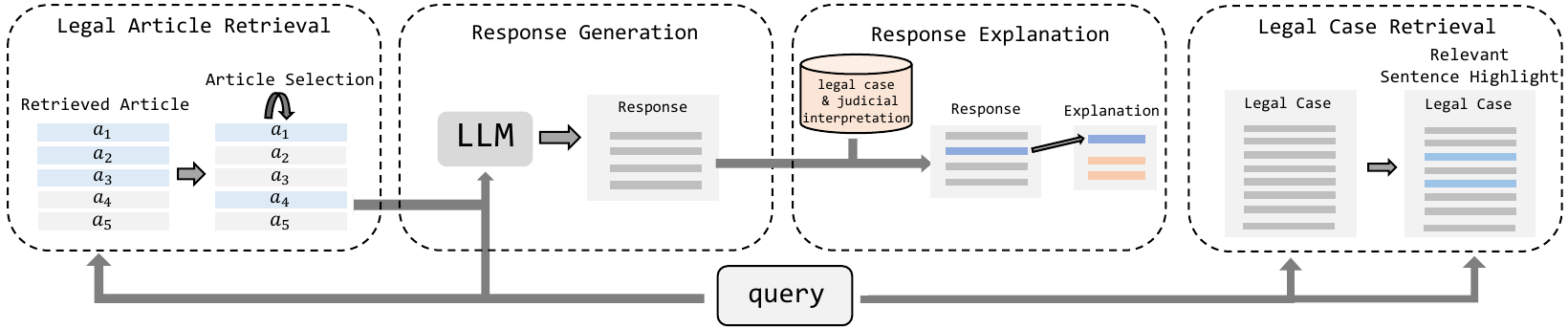}
    \caption{The system architecture overview.}
    \label{fig:system}
\end{figure*}

In this section, we detail the implementation of all back-end models of ELLA.

\subsection{Legal Consultation}
In our work, we use Lawyer LLaMA~\cite{lawyer-llama-report}, a LLM adapted to the legal domain, for legal consultation. Based on Lawyer LLaMA, which focuses on answering queries about marriage, ELLA mainly provides marriage consultation services for users. Since our back-end model is pluggable, we can also replace Lawyer LLaMA with other legal domain LLMs, such as DISC-LawLLM~\cite{yue2023disclawllm}, ChatLaw~\cite{cui2023chatlaw} or LawGPT~\cite{nguyen2023brief}.

\subsection{Legal Article Retrieval}
We use the legal article retrieval model provided by Lawyer LLaMA. Following Lawyer LLaMA, after the user inputs a query, we retrieve the relevant legal articles, and append the top 3 legal articles to the user's query to generate the response. Besides, we display the top $K_1=10$ retrieved legal articles on the front end. If the user selects some relevant legal articles and requires a new response, in the back end, we append all selected legal articles to the input prompt, and LLM will generate a new response.

\subsection{Response Interpretation}

The response interpretation module aims to provide the legal article basis and judicial interpretations for each sentence of the response from the LLM. Here, we use BGE~\cite{bge_embedding}, a state-of-the-art embedding model for retrieval augmented generation. Since BGE has only been pre-trained on the general corpus, it lacks knowledge about the legal domain, thus being unable to distinguish between two terminologies that are semantically similar but have different definitions in the legal domain. Therefore, we need to fine-tune BGE with legal corpus to make it learn legal knowledge.

Due to the lack of training data, we construct a dataset for response interpretation. We sample 2k queries from the legal instruction tuning data published by Lawyer LLaMA. For each query $q$, we obtain the top 3 relevant articles $[a_1, a_2, a_3]$ with the legal article retrieval module, and append these three laws individually to $q_i$. Then Lawyer LLaMA generates different responses $[r_1, r_2, r_3]$ based on the different legal articles. For $r_i = [s_{i1}, s_{i2}, ..., s_{in}]$, we calculated the similarity between each sentence $s_{ij}, j \in [1, n]$ and $a_i$ using BM25~\cite{10.1561/1500000019}. As illustrated in Figure~\ref{fig:dataset}, we treat the sentence with the highest BM25 score $s_{ik}$ and gold article $a_i$ as the positive case $(s_{ik}, a_i)$, while the two most irrelevant sentences $s_{ix}$, $s_{iy}$ as negative cases $(s_{ix}, a_i)$ and $(s_{iy}, a_i)$. We also created negative cases $(s_{ik}, a_t), t \in [1, 2, 3]$ and $t \neq i$ for distinguishing relevant sentence in $r_i$ from other retrieved legal articles. 

Given the similar language style and content between legal articles and judicial interpretations, and the fact that legal articles contain all the terminologies involved in judicial interpretations, we only used legal articles to construct the dataset. After fine-tuning the BGE on this dataset, we obtained a new model, which we denote as $BGE_{1}$ here.

During inference, we use BGE$_1$ to calculate the cosine similarity between the embedding of each sentence in the response and the legal articles and judicial interpretations. If the similarity exceeds a threshold $Thr_1$, we think the corresponding legal article or judicial interpretation can explain the sentence. $Thr_1$ is a hyper-parameter, which we set as 0.85 in our work. Then, we return the articles and judicial interpretations referenced by each sentence to the front end, to help users better understand the LLM's responses.

\subsection{Legal Case Retrieval}
\label{sec:lecard}
In this module, we first retrieve relevant legal cases based on the user's input. Then we find all the key sentences in the legal case that are related to the consultation query. Finally, we re-rank the top $K_2$ retrieved legal cases according to the number of relevant sentences in the legal case, and return the top $K_3$ re-ranked legal cases to the front end.

{\bf Legal Case Retrieval.} Similarly, due to the lack of relevant legal domain knowledge in BGE, we need to fine-tune BGE with the legal domain corpus. Here, we use the dataset LeCaRD~\cite{ma2021lecard}, a publicly available Chinese legal case retrieval dataset. We allocated 80\% of LeCaRD as the training set and 10\% each as the validation and test set. We fine-tune BGE on the training set. Here we denoted the fine-tuned BGE as BGE$_2$. When the user inputs a query, we use BGE$_2$ to retrieve relevant legal cases.

{\bf Relevant Sentence Highlight.} We use BGE$_2$ to calculate the similarity between the user's query and each sentence in the legal case. When the cosine similarity score is larger than $Thr_2$, we consider this sentence to be related to the user's query, thus this sentence can serve as a reason for this case being a relevant legal case. $Thr_2$ is a hyper-parameter, which we set to 0.65 in our work. We highlight all relevant sentences in the case for users, helping them quickly locate the parts of the case that are highly related to their query. In this way, users can quickly judge whether this legal case is relevant and helpful, and they can also quickly obtain important information that they care about.

{\bf Legal Case Re-rank.} We think that the more relevant sentences in a case, the larger the possibility of the case being a relevant legal case. Therefore, we re-rank the top $K_2$ legal cases retrieved by BGE$_2$ according to the number of relevant sentences, and return the re-ranked top $K_3$ legal cases to the front end. We set $K_2 = 50$ and $K_3 = 15$ in our work.

\section{Evaluation}
In this section, we automatically evaluate our case retrieval model. We also conduct a user study to evaluate whether ELLA helps users obtain more accurate, interpretable and informative information during the consultation.

\subsection{Automatical Evaluation}
\begin{table}[t]
    \centering
    \small
    \resizebox{0.48\textwidth}{!}{
    \begin{tabular}{c|c|c|c}
        \toprule
        Model& NDCG@10 & NDCG@20 & NDCG@30 \\
        \midrule
        BM25 & 53.51 & 55.81 & 58.03 \\
        BGE & 66.57 & 67.13 & 71.91 \\
        BGE$_2$ & 76.34 & 77.84 & 78.29 \\
        CaseEncoder~\cite{ma2023caseencoder} & 78.5 & 80.3 & 83.9 \\
        SAILER~\cite{li2023sailer} & 79.79 & 82.26 & 84.85 \\
        CaseFormer~\cite{su2024caseformer} & 83.45 & 83.57 & 83.94 \\
        \bottomrule
    \end{tabular}
    }
    
    \caption{Results of Legal Case Retrieval Model.}
    \label{table:case_retrieval}
\end{table}

As we mentioned in Section~\ref{sec:lecard}, we split LeCaRD into 80\% for training, 10\% for validation and 10\% for testing. Here, we use the LeCaRD test set to evaluate our legal case retrieval model, BGE$_2$. Following CaseEncoder~\cite{ma2023caseencoder}, we use the Normalize Discounted Cumulative Gain (NDCG) metric as the evaluation metric. The experimental results are shown in Table~\ref{table:case_retrieval}. 

Compared with BM25 and BGE which has not been fine-tuned, BGE$_2$ shows a significant increase in each NDCG@K. This shows that the fine-tuned BGE can learn legal knowledge well, and better distinguish legal cases that are semantically similar but not relevant in the legal domain. Although CaseEncoder~\cite{ma2023caseencoder}, SAILER~\cite{li2023sailer} and CaseFormer~\cite{su2024caseformer} outperform BGE$_2$, we use BGE$_2$ since it can serve as an embedding model for relevant sentences similarity matching mentioned in Section~\ref{sec:lecard}. Note that our legal case retrieval model is pluggable, so we can also additionally add SOTA models mentioned above for legal case retrieval.

\subsection{User Study}

\subsubsection{Study Design}

We conduct a user study to validate whether ELLA can improve users' legal consultation experience. Since LLMs deliver an impressive performance in answering simple questions, such as {\it "Can I get married if I am younger than 20?"}, we randomly selected 20 consultation queries about complex marriage situations for the user study. We invited 3 non-legal professional users and asked them to obtain solutions to these queries through ELLA. Users will evaluate whether the three modules in ELLA are helpful for their legal consultation.

\subsubsection{Result}
\paragraph{Response Regeneration.} For an average of 83\% of the queries, users find that the top 3 legal articles retrieved are not entirely correct, impeding LLM from directly generating correct responses based on these articles. For an estimated 20\% of the queries, LLMs can not provide correct responses due to the noise brought by irrelevant legal articles, while for 25\%, LLM's responses are incomplete, as relevant legal articles were not among the initial top three results. Another 38\% of responses contained irrelevant information resulting from the inclusion of unrelated legal articles within the top three results. However, in 80\% cases, users can successfully receive correct responses by selecting relevant legal articles for LLM to regenerate responses.

\paragraph{Response Interpretation.} Users have reported that for approximately 95\% of the queries, ELLA can accurately provide the legal article basis of the responses generated by the LLM. By cross-referencing the responses with the corresponding legal article, users can swiftly determine whether the responses are reliable or inaccurate. For instance, when a user asks, {\it "I have never had children since I got married, and now I am planning to adopt a child from a relative. Can I adopt a child privately?"} LLM responds {\it "Adopters need to meet the following conditions..."}. ELLA justifies the response by citing Article 1098 of the Civil Code as its legal article basis. Additionally, it retrieves Article 1100 of the Civil Code, {\it "A childless adopter may adopt two children...,"} which the user can select for the LLM to generate a full response. Users also noted that, in about 73\% of the queries, parts of the legal articles have already been included within the responses. However, LLM may not fully rephrase the entire article. By providing the legal articles basis, users can conveniently access to the complete information in the legal article.

In all provided judicial interpretations, roughly 30\% serve the purpose of clarifying specific legal terminologies or special cases. For instance, consider a scenario where a user inquires, {\it "My husband and I have obtained a marriage certificate but have not cohabited. We are now filing for divorce and my husband wishes to return the bride price. Is this permissible?"} In response, ELLA gives additional judicial interpretation that illuminates the conditions under which the return of the bride price is allowed. However, for the remaining 70\%, users claim that they are already familiar with the content in the judicial interpretations, such as, {\it "Support payments encompass children's living expenses, education costs, medical bills and other expenditures."} Generally speaking, users assert that judicial interpretations can assist them in acquiring a better comprehension of the responses when interpretation is required, facilitate accurate judgments according to their situations, and pave the way for further consultation tailored to the specifics of their current circumstances.

\paragraph{Legal Case Retrieval.} On average, 77\% of queries proved the legal case retrieval module to be beneficial for user consultations. Users conveyed that although the retrieved legal cases might not exactly match their situations, these cases provide a reference point to gauge the possible outcomes for their unique circumstances. All users concurred that highlighting pertinent sentences significantly streamlines the process of reading cases. By emphasizing the information users are interested in, the user's reading efficiency improves.

\section{Conclusion}
We present a novel tool, ELLA, for legal consultation. ELLA provides the legal basis and judicial interpretations that supplement the legal advice generated by LLMs, increasing users' understanding and trust in LLM responses. It also displays retrieval results from the retrieval model and allows users to actively select relevant legal articles, thereby assisting the LLMs in generating more accurate responses. Additionally, equipped with a legal case retrieval model, users can refer to relevant legal cases for more comprehensive information. ELLA enables LLMs to provide legal advice that is easier to interpret, more precise, and more informative.

\section*{Acknowledgments}
This work is supported in part by NSFC
(62161160339) and Beijing Science and Technology Program (Z231100007423011).
We thank the anonymous reviewers for their valuable comments and suggestions.
For any correspondence, please contact Yansong Feng.

\section*{Limitations}
For simple legal queries, legal LLMs can provide correct responses in most cases. ELLA primarily assists with complex legal consultation queries. When users ask multiple questions within a single input, our legal article retrieval module may not comprehensively extract all relevant legal articles. In future work, we plan to integrate different retrieval modules to increase the diversity of retrieved legal articles. 

As official judicial interpretations only contain 76 articles, ELLA can not provide interpretations for all professional terminologies. We will incorporate additional external legal knowledge, such as legal textbooks, to provide interpretations for more professional terminologies. 

Due to limited computational resources, we do not use state-of-the-art case retrieval modules. We will adopt them in our future work.

\section*{Ethics Statement}
The main purpose of this paper is to explore how to provide users with better legal consultation services. However, it is important to note that the outputs generated by the model may contain non-standard, incorrect, gender-biased, or morally questionable information. Therefore, please adopt the legal advice provided by the model with caution. When legal assistance is needed, please seek help from qualified professionals.

The external knowledge used in this paper, such as legal articles and legal cases, was obtained from the official websites of the Chinese government. The data does not contain any private information and cannot be used for commercial purposes.

All participants involved in the user study are voluntary and anonymous. We did not collect any private information from the participants.

\bibliography{custom}
\bibliographystyle{acl_natbib}

\appendix
\clearpage

\section{Dataset Construction}
\begin{figure}[h]
    \center
    \includegraphics[width=0.408\textwidth]{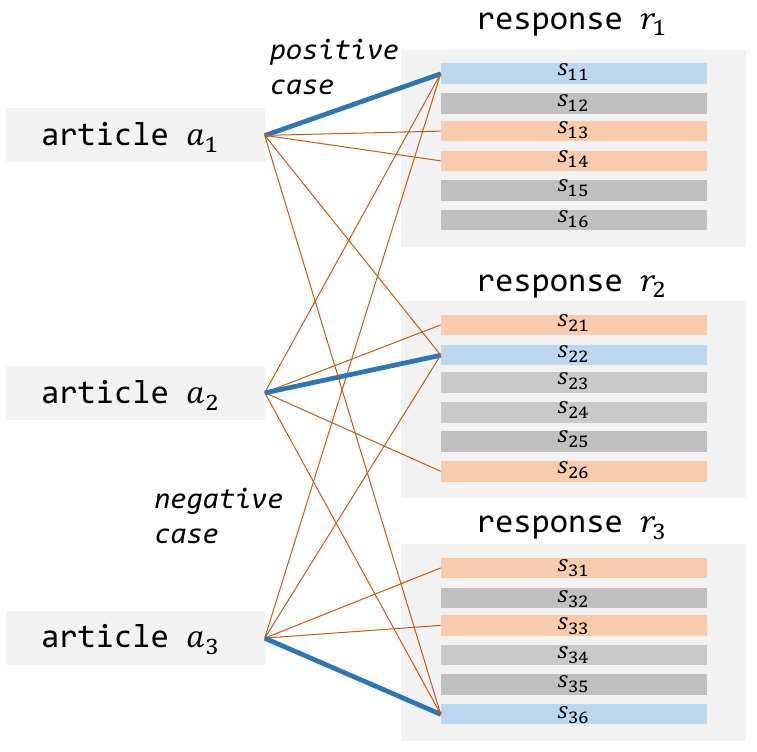}
    \caption{Schematic of Dataset Construction. The blue sentences indicate the sentences with the highest BM25 scores and the orange sentences are the most irrelevant ones. Blue lines indicate positive cases and orange lines indicate negative cases.}
    \label{fig:dataset}
\end{figure}

\section{Consultation Examples}
\label{app:conversation}


\begin{table*}[b]
  \centering{
  \begin{tabular}{p{8cm}|p{8cm}}
    \toprule
    \begin{CJK*}{UTF8}{gbsn}$Q_1:$ 收养小孩有什么要求吗？\end{CJK*} & $Q_1:$ What are the requirements for adopting a child? \\
    \midrule
    \begin{CJK*}{UTF8}{gbsn}$R_1:$ 您好，根据《民法典》的规定，收养人需要同时具备以下五个条件：1.无子女或只有一名子女；2.有抚养、教育和保护被收养人的能力；3.未患有在医学上认为不应当收养子女的疾病；4.无不利于被收养人健康成长的违法犯罪记录；5.年满三十周岁。此外，如果收养人有配偶，则需要夫妻共同收养。收养关系自登记之日起成立，收养关系当事人自愿签订收养协议的，可以签订收养协议。如果收养关系当事人各方或一方要求办理收养公证的，应当办理收养公证。县级以上人民政府民政部门应当依法进行收养评估。请问您还有其他方面的问题吗？\end{CJK*} & $R_1:$ Hello, according to the provisions of the Civil Code, adopters need to meet the following five conditions simultaneously: 1. Having no children or having only one child; 2. Having the ability to support, educate, and protect the adopted child; 3. Not suffering from diseases that are considered by medical standards as reasons not to adopt children; 4. Having no criminal records that are detrimental to the healthy growth of the adopted child; 5. Being at least thirty years old. Additionally, if the adopter is married, the adoption should be done jointly by the couple. The adoption relationship is established from the date of registration. If the parties to the adoption voluntarily sign an adoption agreement, they can do so. If the parties to the adoption or one party requests adoption notarization, adoption notarization should be processed. The civil affairs department of the people's government at or above the county level shall conduct adoption assessments in accordance with the law. Do you have any other questions? \\
    \midrule
    \begin{CJK*}{UTF8}{gbsn}$R_1':$ 您好，根据《民法典》的规定，收养小孩需要满足以下条件：1.无子女的收养人可以收养两名子女；2.有子女的收养人只能收养一名子女；3.收养孤儿、残疾未成年人或者儿童福利机构抚养的查找不到生父母的未成年人，可以不受前款和本法第一千零九十八条第一项规定的限制。此外，无配偶者收养异性子女的，收养人与被收养人的年龄应当相差四十周岁以上。最后，符合以下条件的未成年人可以被收养：1.丧失父母的孤儿；2.查找不到生父母的未成年人；3.生父母有特殊困难无力抚养的子女。请问您的具体情况是什么？是否符合上述规定呢？\end{CJK*} & $R_1':$ Hello, according to the provisions of the Civil Code, adopting a child requires meeting the following conditions: 1. An adopter with no children can adopt up to two children; 2. An adopter with children can only adopt one child; 3. Orphans, disabled minors, or minors raised by child welfare institutions whose birth parents cannot be found are not subject to the restrictions in the preceding paragraph and Article 1098 of this Law. Additionally, if an unmarried person adopts a child of the opposite sex, the adopter's age must be at least forty years older than the adoptee's. Lastly, the following minors can be adopted: 1. Orphans who have lost both parents; 2. Minors whose birth parents cannot be found; 3. Children whose birth parents face special difficulties and are unable to raise them. Could you please provide specific details about your situation? Do you meet the above requirements? \\
    \bottomrule
  \end{tabular}
    }
    \caption{$R_1$ is generated based on Article 1098, 1101, and 1105 of Civil Code. $R_1'$ is generated based on Article 1100, 1102, and 1093 of Civil Code.}
    \label{table:q1}
\end{table*}

\clearpage

\begin{table*}[htbp]
  \centering
  \begin{tabular}{p{8cm}|p{8cm}}
    \toprule
    \begin{CJK*}{UTF8}{gbsn}$Q_2:$ 我的情况是这样的，我女朋友结婚了，但是已经和她老公分居好几年，是不是算自动离婚了?我们没以夫妻的名义同居算重婚吗? \end{CJK*} & $Q_2:$ Here's my situation: my girlfriend got married, but she has been separated from her husband for several years. Does this count as automatic divorce? Would living together without being married constitute bigamy? \\
    \midrule
    \begin{CJK*}{UTF8}{gbsn}$R_2:$ 您好，根据我国婚姻法规定，夫妻双方自愿离婚，可以自行协商处理财产分割等问题，不需要经过法院审理，如果您女友与她的丈夫已经分居多年，并且没有其他原因导致他们无法继续维持婚姻关系，那么可以视为自动离婚。至于您们是否构成重婚，需要了解你们是否已正式离婚，以及是否已经与他人结婚或同居。如果已经离婚，则不存在重婚的问题。如果还没有离婚，则需要遵守相关法律法规，如禁止重婚等。请问您们是否已经离婚?\end{CJK*} &$R_2:$ Hello, according to the marriage laws of our country, if both parties voluntarily divorce, they can negotiate property division and other issues themselves without the need for court trial. If your girlfriend has been separated from her husband for many years and there are no other reasons preventing them from continuing their marriage, it can be considered an automatic divorce. As for whether you constitute bigamy, it depends on whether you have officially divorced and whether you have married or cohabited with someone else. If you have already divorced, there is no issue of bigamy. If not, you need to comply with relevant laws and regulations, such as the prohibition of bigamy. May I ask if you have already divorced? \\
    \midrule
    \begin{CJK*}{UTF8}{gbsn}$R_2':$ 您好，根据我国法律规定，只有在完成离婚登记或者离婚判决书、调解书生效后，才能够真正意义上解除婚姻关系。而您所提到的情况，虽然你们没有以夫妻的名义同居，但并不意味着已经完成了离婚手续。因此，您们目前仍然处于婚姻状态之中。另外，重婚是指在未办理离婚手续的情况下再次结婚的行为，而您所提到的情况并不属于重婚范畴之内。请问您还有其他方面的问题需要解答吗?\end{CJK*} & $Q_2':$ Hello, according to the laws of our country, marriage is only truly dissolved after completing divorce registration, or when the divorce judgment or mediation agreement takes effect. The situation you mentioned, although you haven't lived together as spouses, does not mean that the divorce procedures have been completed. Therefore, you are still in a marital status currently. Additionally, bigamy refers to remarrying without completing divorce procedures, which doesn't apply to the situation you mentioned. Do you have any other questions that need answering? \\
    
    \bottomrule
  \end{tabular}
  \caption{$R_2$ is generated based on Article 1080, 1091, and 1043 of Civil Code. $R_2'$ is generated based on Article 1080 of Civil Code.}
\end{table*}

\clearpage
\begin{table*}[htbp]
  \centering
  \begin{tabular}{p{8cm}|p{8cm}}
  \toprule
  \begin{CJK*}{UTF8}{gbsn}$Q_3:$ 我的情况是这样的，我女朋友结婚了，但是已经和她老公分居好几年，是不是算自动离婚了?我们没以夫妻的名义同居算重婚吗?\end{CJK*} & $Q_3:$ My situation is like this: My girlfriend got married, but she has been separated from her husband for several years. Does this count as automatic divorce? Would living together without being married constitute bigamy? \\
  \midrule
  \begin{CJK*}{UTF8}{gbsn}$R_3:$ 您好，根据《民法典》的规定，只有完成离婚登记或者离婚判决书、调解书生效才能够解除婚姻关系。而您的女友已经与她的丈夫分居多年，但并未办理离婚手续，因此不能算自动离婚。同时，如果您们没有以夫妻的名义同居，则不属于重婚行为。不过，在婚姻关系中，夫妻双方应该互相忠实、互相尊重、互相关爱，并且要保持良好的家庭环境。如果出现家庭矛盾或纠纷，可以通过协商解决，也可以通过司法途径解决。请问您是否需要进一步了解如何处理此类问题呢?
\end{CJK*} & $R_3:$ Hello, according to the provisions of the Civil Code, marriage can only be dissolved by completing divorce registration or when a divorce judgment or mediation agreement becomes effective. As your girlfriend has been separated from her husband for several years without completing divorce procedures, it does not count as automatic divorce. Also, if you haven't lived together as husband and wife, it does not constitute bigamy. However, in a marital relationship, both parties should be faithful, respect each other, love each other, and maintain a good family environment. If there are family conflicts or disputes, they can be resolved through negotiation or legal means. Would you like further information on how to handle such issues? \\

  \bottomrule
  \end{tabular}
  \caption{$R_3$ is generated based on Article 1080, 1091, 1043 of Civil Code.}
\end{table*}
\clearpage

\begin{table*}[htbp]
  \centering
  \begin{tabular}{p{8cm}|p{8cm}}
    \toprule
    \begin{CJK*}{UTF8}{gbsn}$Q_4:$ 我的情况是这样的，我女朋友结婚了，但是已经和她老公分居两年多了，算不算自动离婚了？我们没以夫妻的名义同居算重婚罪吗？\end{CJK*} & Here's my situation: my girlfriend got married, but she has been separated from her husband for over two years. Does this count as automatic divorce? Would it be considered bigamy if we cohabitated without being married? \\
    \midrule
    \begin{CJK*}{UTF8}{gbsn}$R_4:$ 您好，根据我国婚姻法规定，夫妻双方自愿离婚，可以自行协商离婚事宜并到当地民政局办理离婚手续。如果您女友与她的丈夫没有达成协议，则需要通过法院诉讼程序来解决离婚问题。至于您提到的同居问题，如果两人没有以夫妻的名义同居，则不会构成重婚罪。不过，建议您们先了解清楚相关法律规定，再做出决定。另外，关于收养小孩的问题，您需要提供更多信息，例如孩子的性别、年龄等，才能够更好地为您解答问题。\end{CJK*} & Hello, according to the Marriage Law of our country, if both parties in a marriage voluntarily agree to divorce, they can negotiate the terms of the divorce and go to the local civil affairs bureau to handle the divorce procedures. If your girlfriend and her husband have not reached an agreement, they will need to resolve the divorce issue through the court litigation process. As for the cohabitation issue you mentioned, it would not constitute bigamy if you two did not cohabit under the guise of marriage. However, it's advisable for you to understand the relevant legal provisions before making any decisions. Additionally, regarding the adoption of a child, you would need to provide more information, such as the child's gender and age, for me to better assist you with your questions. \\

    \bottomrule
  \end{tabular}
  \caption{$R_4$ is generated based on Article 1080, 1091, 1043 of Civil Code.}
    \label{table:q4}
\end{table*}

  
  

\end{document}